\documentclass[runningheads]{llncs}
\usepackage[T1]{fontenc}
\usepackage{booktabs}
\usepackage{listings}
\usepackage{subcaption}
\usepackage{graphicx}
\usepackage{wrapfig}

\begin{document}
\title{Exploring Modularity of Agentic Systems for Drug Discovery}

\author{Laura van Weesep\inst{1,2}\orcidID{0009-0009-5524-7960}\and\\
Samuel Genheden\inst{1}\orcidID{0000-0002-7624-7363} \\
Ola Engkvist\inst{1,3}\orcidID{0000-0003-4970-6461} 
\\Jens Sjölund \inst{2}\orcidID{0000-0002-9099-3522}}

\authorrunning{van Weesep et al.}

\institute{Molecular AI, Discovery Sciences, R\&D, AstraZeneca, Gothenburg, Sweden \email{\{lauradesire.vanweesep,samuel.genheden,ola.engkvist\}@astrazeneca.com}\and
Department of Information Technology, Uppsala University, Sweden \and Department of Computer Science and Engineering, Chalmers University of Technology and University of Gothenburg, Sweden}

\maketitle     
\begin{abstract}
Large-language models (LLMs) and agentic systems present exciting opportunities to accelerate drug discovery. In this study, we examine the modularity of LLM-based agentic systems for drug discovery, i.e., whether parts of the system such as the LLM and type of agent are interchangeable, a topic that has received limited attention in drug discovery. We compare the performance of different LLMs and the effectiveness of tool-calling agents versus code-generating agents.  Our case study, comparing performance in orchestrating tools for chemistry and drug discovery using an LLM-as-a-judge score, shows that Claude-3.5-Sonnet, Claude-3.7-Sonnet and GPT-4o outperform alternative language models such as Llama-3.1-8B, Llama-3.1-70B, GPT-3.5-Turbo, and Nova-Micro. Although we confirm that code-generating agents outperform the tool-calling ones on average, we show that this is highly question- and model-dependent. Furthermore, the impact of replacing system prompts is dependent on the question and model, underscoring that even in this particular domain one cannot just replace components of the system without re-engineering. Our study highlights the necessity of further research into the modularity of agentic systems to enable the development of reliable and modular solutions for real-world problems.

\end{abstract}
\section{Introduction}
Drug discovery is a complex, multidisciplinary field \cite{szymanski2011adaptation} that has increasingly integrated artificial intelligence (AI) over the past several decades \cite{serrano2024artificial}.
The next step in AI for drug discovery is anticipated to be the development of research collaborators, which no longer rely on humans to specify each step and manually transfer the output from one tool to the other, but think independently, automate workflows, and ask for input when needed. 
LLM-based agents that use cheminformatic tools are starting to bring this to reality \cite{gao2025pharmagents,inoue2024drugagent,kim2025mt}. 

\begin{wrapfigure}{r}{0.4\textwidth}
  \begin{center}
  \vspace{-10mm}
    \includegraphics[width=\linewidth,trim={6mm 15mm 6mm 1mm}, clip]{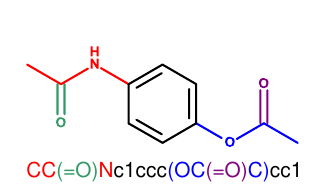}
  \end{center}
  \vspace{-5mm}
    \caption{Molecule with its SMILES notation. Brackets correspond to branches, \textbf{=} to a double bond.}
        \label{fig:SMILES}
        \vspace{-7mm}
\end{wrapfigure}

The progress in LLMs and their reasoning has been remarkable \cite{Zhang2025-nj}. While the reasoning frameworks enhance logical coherence, they lack chemical intuition \cite{cui2025can}. Chemical language is often challenging for LLMs \cite{jang2025improving}, especially in SMILES (Simplified Molecular Input Line Entry System) \cite{weininger1988smiles} strings, which encode molecular structures as text. Reasons for this include that molecules are more often described with their common name or IUPAC name than with SMILES strings, resulting in relatively few SMILES strings in the training data of most foundational LLMs \cite{klinger2008detection}. Additionally, atoms close in the molecular structure are not necessarily close to each other in the text representation, making it more challenging for the LLM to navigate the structure of the molecule (see Figure \ref{fig:SMILES}). Importantly, language models do not have an inherent understanding of SMILES and instead treat them as ordinary characters, which adds to the challenge \cite{guo2023can}. 
To mitigate the shortcomings of LLMs, researchers have begun to augment LLMs with specialized chemistry tools. 

Agentic AI, a type of AI system that independently makes decisions and interacts with its surroundings \cite{hosseini2025role}, has already been experimented with in discovery sciences. Agentic AI could reduce human errors, enable non-technical wet-lab scientists to harness computational tools, and speed up research workflows significantly. For example, ChemCrow \cite{m2024augmenting} integrates GPT-4 with external chemical tools using the LangChain framework\footnote{https://github.com/langchain-ai/langchain}, enabling it to perform practical chemistry tasks like synthesis planning, molecule generation, and property prediction. PharmAgents \cite{gao2025pharmagents} is more focused on the pharmaceutical industry, simulating the full drug discovery workflow. 
  Other examples include CoScientist \cite{boiko2023autonomous} for chemical reaction discovery, DrugAgent for drug discovery tasks such as property prediction \cite{inoue2024drugagent} and Robin for semi-automatic discovery of therapeutic candidates \cite{ghareeb2025robin}. Although they achieve impressive performance by tailoring the tool orchestration and prompt engineering for this use case, they have not elaborately covered the workflow's adaptability to new LLMs and tools. Some works do claim that the LLM is replaceable, yet do not elaborate on the effect of doing this \cite{grougan2025librarian}. 
  There have been recent benchmarks published on the capabilities of LLMs in the chemical domain \cite{cai2025mollangbench,guo2023can,mirza2025framework,runcie2025assessing}, yet they focus on the LLM not agentic AI, and, even if they use tools, often do not study the use of those tools themselves. Additionally, capabilities are often summarized over a dataset, without considering why some questions are answered more accurately than others, despite research highlighting that the type of question and even the phrasing of the question influence the competitiveness of LLMs \cite{guo2023can,mirza2025framework}. Besides, LLMs are frequently finetuned to enhance performance \cite{hu2025aitomia,li2025drugpilot,wang2025txgemma,ye2025drugassist}, which means that LLMs cannot just be replaced without retraining. 

To benefit from the developments both in the field of foundational LLM models and the continuous improvement of cheminformatics tools, these agentic systems should be modular, allowing tools and backbone LLMs to be swapped and prompt improvements to lead to consistent improvement independent of the model. However, the prevailing focus on performance has (so far) overshadowed concerns about modularity. 

In this paper, we examine the modularity of LLM-based agentic systems in drug discovery. One of our key contributions is to compare seven LLMs, revealing that the agentic system's performance is highly dependent on the underlying LLM. We further compare the JSON-based tool calling (ToolCallingAgent) and code generation (CodeAgent) agent type, also zooming in on the performance on a question-specific level. Finally, we test whether prompt engineering strategies, such as providing clearer examples or domain-specific instructions, yield improvements across different LLMs and questions. Additionally, we employ the lightweight, Hugging Face-based smolagents framework and demonstrate its application in a drug discovery context for the first time.

\section{Methods}
In this section, we cover the agentic framework, the specific types of agents used, and the experimental setup. This includes a description of the smolagents framework and the LLM-as-a-judge used to automatically assess the different agents in our comparison. 

\subsection{Agentic Framework}
We use the recently released smolagents framework\footnote{https://github.com/huggingface/smolagents}, which intends to be more lightweight and flexible compared to alternatives such as LangChain\footnote{https://www.langchain.com/}. The smolagents framework supports multiple LLMs for answering user queries, including those hosted via Azure OpenAI and integrated through LiteLLM\footnote{ https://huggingface.co/docs/smolagents/}. Agents in the smolagents framework are multi-step agents that use a ReAct-inspired system framework \cite{yao2023react} in which actions are carried out by first reasoning and then acting and feeding the execution output to the next step.

The smolagents framework offers different types of agents. The CodeAgent is capable of directly writing and executing code and in this way avoids having to change the environment and waiting for the output of a preceding tool call. Previous research \cite{wang2024executable} shows that the CodeAgent is more flexible and token-efficient than the JSON-based ToolCallingAgent. The ToolCallingAgent is suited for more rigid workflows where the tool calls and outcomes are more predictable and less creativity is required. 
This study sets out to explore the claimed benefits of both the smolagents framework and CodeAgents for orchestrating cheminformatics tools. 

\subsection{Experimental Setup}

\begin{figure}[t]
    \centering
    \includegraphics[width=0.85\linewidth]{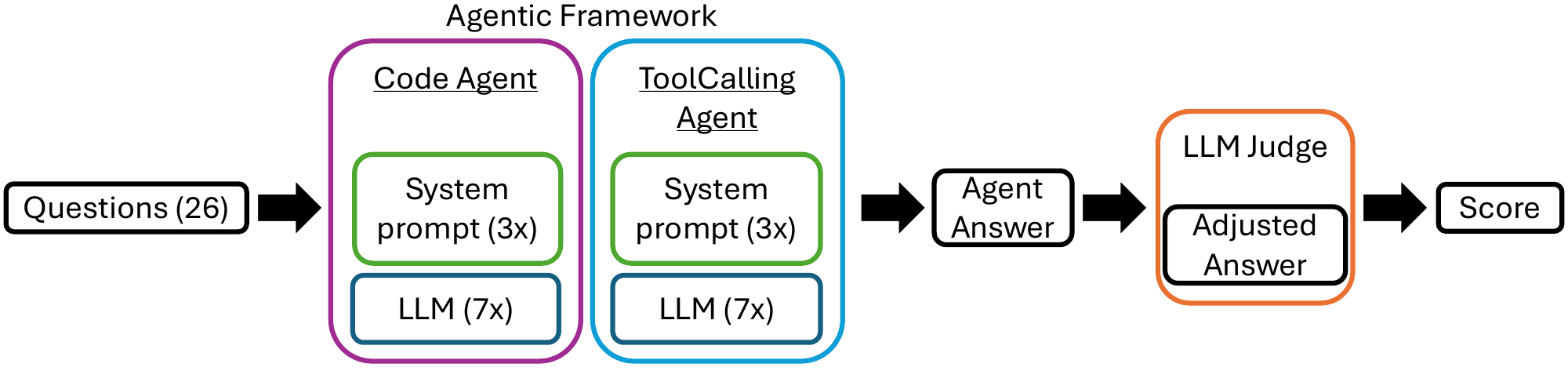}
    \caption{Overview of our experimental setup. We experiment with seven different LLMs, three different system prompts and use both the CodeAgent and the ToolCallingAgent. Ultimately, the performance is assessed using an LLM-as-a-judge.\looseness=-1}
    \label{fig:Experimental_Overview}
    \vspace{-3mm}
\end{figure}

Figure \ref{fig:Experimental_Overview} presents an overview of the experiments done. First, we compare the performance of seven different LLMs, then the CodeAgent with the ToolCallingAgent, and lastly the study influence of adapting the system prompt.

To test the tool orchestration capabilities of different agents we formulated 26 industry-representative cheminformatics questions (see Appendix A.1). 
We consider 17 relatively simple tools (full list in Appendix A.2), including functions from the RDkit package \cite{landrum2013rdkit} and the pubchempy package\footnote{https://pubchempy.readthedocs.io/en/latest/}. This allows for integration with the PubChem website \cite{kim2025pubchem} and gives us tools related to physicochemical properties such as the CrippenLogP and molecular weight. Importantly, we also include tools related to converting text into molecular objects and understanding the attributes of molecular objects. These tools are already used to answer industry-relevant questions manually and allow us to formulate multiple valid chains of reasoning ourselves. All LLMs had access to these tools. 

In total seven different LLM models were experimented with, including GPT-3.5-Turbo, GPT-4o, Claude-3.5-Sonnet, Claude-3.7-Sonnet, Llama-3.1-8B, Llama-3.1-70B and Nova-Micro (see Appendix A.3 for further details). For all models we experimented with both the ToolCallingAgent and the CodeAgent, yet for the Llama models, the ToolCallingAgent gave an error (see Appendix A.4). 

Furthermore, to understand the system prompts' effect on performance we also tested three versions of the system prompt (see Appendix A.5 for the default system prompt of the CodeAgent). We modified the “default” prompt by first removing irrelevant examples of fictional tool usage to create a “clean” prompt, and then added computational chemistry context, replacing the general instructions and including relevant examples to get a “relevant” prompt. 

To score the models we use the LLM-as-a-judge approach using GPT4o as the judge. Following recommendations of a recent survey \cite{gu2024survey}, we include repetitions of each experiment and let the LLM model judge each run independently to generate a score between 0 and 100. To mitigate the common problem of length bias \cite{saito2023verbosity} we prompted the LLM judge to summarize the answer given by the agent first and then judge the summarized answer (see Appendix A.6). 

\section{Results and Discussions}
\begin{figure}[t]
    \centering
    \includegraphics[width=0.55\linewidth]{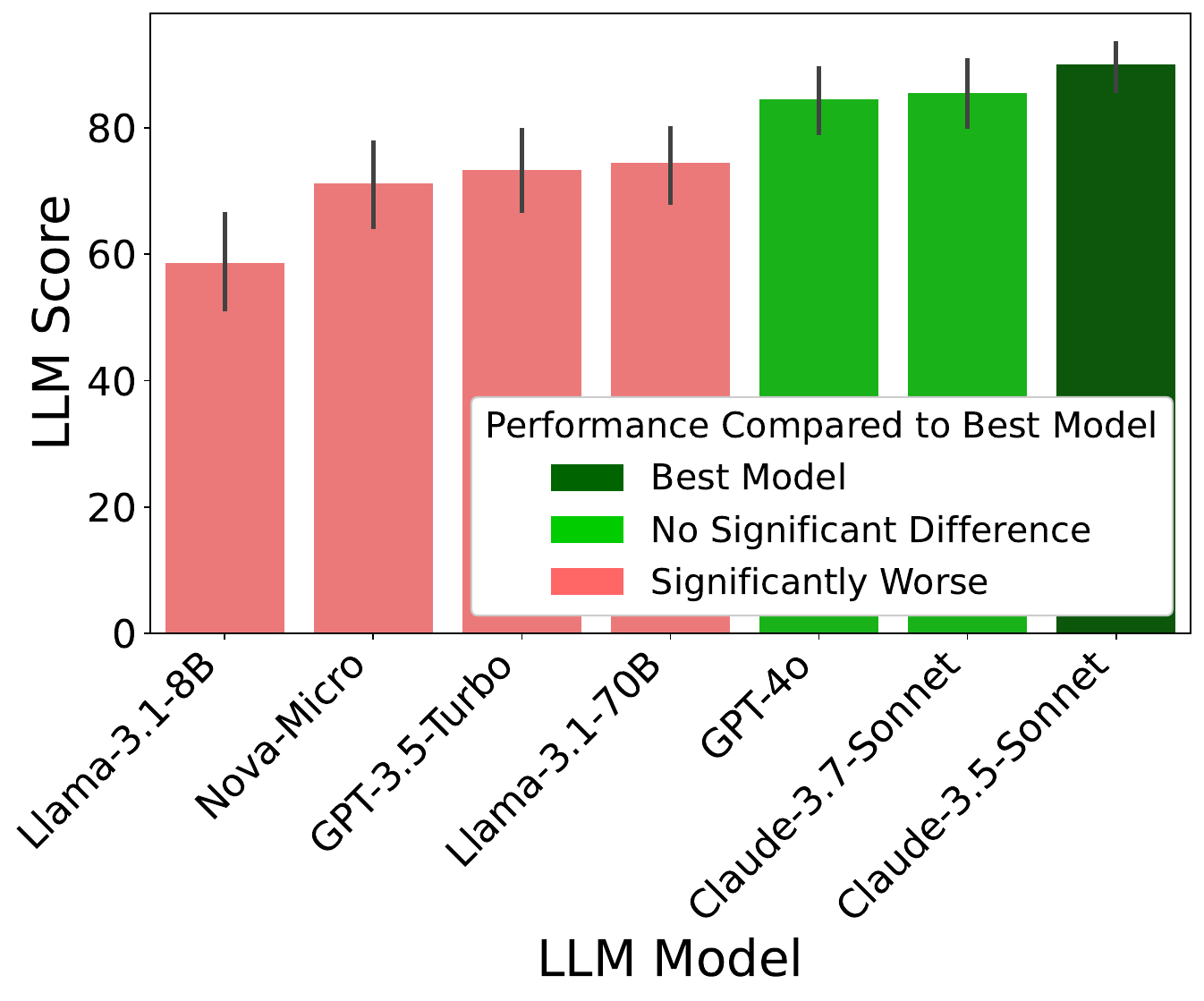}
        \caption{Average with standard deviation LLM score over 26 questions with five repetitions when using the CodeAgent and the default system prompt. Light red indicates a model that performed significantly worse, light green indicates no significant difference and dark green indicates the best model. Bonferroni correction was applied with a significant difference defined as $p<0.05$. \looseness=-2}
    \label{fig:model_file}
    \vspace{-3mm}
\end{figure}
This section presents the experimental results evaluating the effectiveness of different aspects of the agentic framework across the cheminformatics questions. We start by comparing the performance of the different LLMs and continue by studying the performance of the CodeAgent and the ToolCallingAgent. Subsequently, we analyze the impact of prompt engineering on task performance. 

\subsection{Comparing LLMs}

To investigate the difference in performance between LLMs we evaluated the CodeAgent with the smolagents default system prompt for the seven different LLMs. For each LLM the 26 questions were asked five times and independently scored by the LLM-as-a-judge system. As shown in Figure \ref{fig:model_file} Claude-3.5-Sonnet, Claude-3.7-Sonnet and GPT-4o achieve a higher mean score than the other models and lower standard deviation. A paired one-sided Wilcoxon rank test ($p<0.05$) on the mean value over the 5 repeats per question showed that the Llama models, GPT-3.5-Turbo and the Nova-Micro model perform significantly worse than the best model. Claude models have previously been shown to perform well in chemistry \cite{mirza2025framework}. While this may suggest that a better domain understanding in the underlying LLM enhances the tool orchestrating capabilities in that domain, a counterexample is the general-purpose biomedical AI agent Biomini for which this was not observed \cite{huang2025biomni}.

\begin{wrapfigure}[29]{r}{0.408\textwidth}
    \vspace{-1mm} % small tweak if needed
    \centering
    \includegraphics[width=\linewidth]{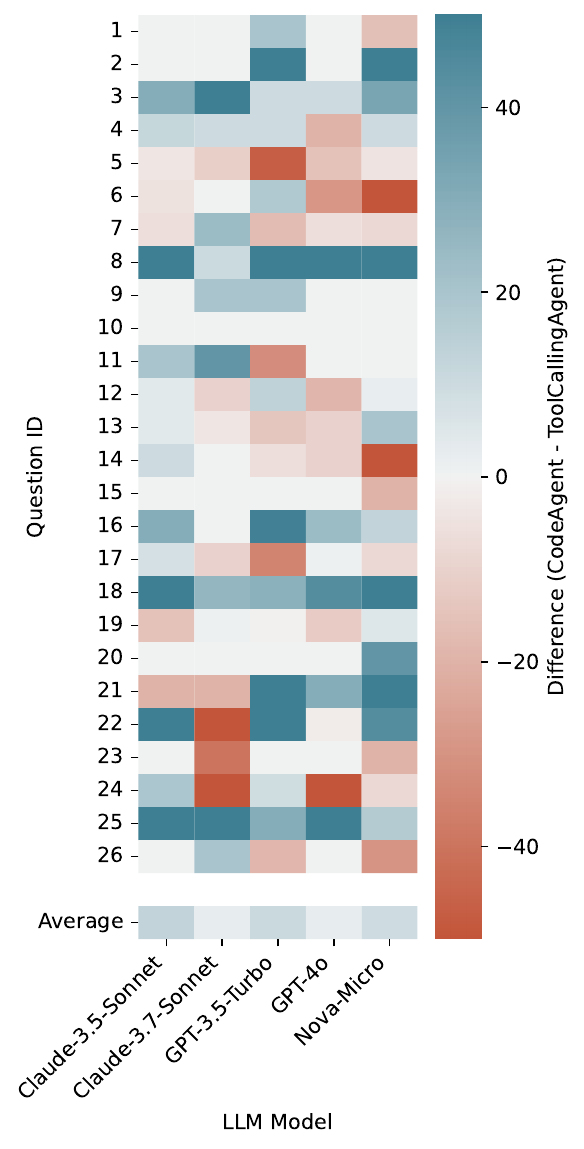}
    \caption{LLM score difference between CodeAgent and ToolCallingAgent across 26 questions and 5 LLMs (average over 5 runs).}
    \label{fig:heatmap_code_vs_tool}
\end{wrapfigure}
Looking beyond average performance, there are a few questions where models that perform worse overall match or even surpass the top-performing ones (see Appendix A.7). All models answer question 20 (``How many rotatable bonds does ethene have?''), correctly for instance. This question has a straightforward answer (``0 rotatable bond''), which could be guessed. However, since ethene is the most basic double-bond molecule, this information is likely included in the training data of the LLM. Upon closer inspection, we indeed found that the tools are not always used to answer this question. Furthermore, we can see that for the more complex compound in the form of a SMILES  in question 21 (``How many rotatable bonds does \texttt{CC(C)C1CCC1} have?'') the performance varies more across LLMs and top models no longer always answer this question correct. Finally, the results on question 8 (``What is the molecular weight of \texttt{C1OCc2c1ccc3S=CC=Cc23?}'') and question 12 (``Does guanine have more hydrogen bond donors or acceptors?'') show other models outperform the Claude models and GPT-4o. This leaves room for further investigations to understand which factors contribute to those specific questions showing different trends across LLMs.

\subsection{ToolCallingAgents versus CodeAgents}

Beyond the reliance on the specific LLM, to corroborate that CodeAgents perform better than  ToolCallingAgents also in our case study, we compare their LLM scores. Figure \ref{fig:heatmap_code_vs_tool} shows the result as a heatmap indicating whether the CodeAgent (blue) or ToolCallingAgent (red) performs better for that combination of model and question on average.

In terms of overall performance, the CodeAgent is better than the ToolCallingAgent in most cases, as indicated by a greater prevalence of blue compared to red and for each of the models, the CodeAgent outperformed the ToolCallingAgent on average (by 2.5 up to 12.9 units). 
However, a closer look reveals that this is also dependent on the question. For instance, on question 3 (``How many hydrogen acceptors does 
\texttt{C1OCc2c1ccc3S=CC=Cc23} have?''), the CodeAgent always performs much better, same for questions 8 and 25 (``What is the molecular weight of \texttt{C1OCc2c1ccc3S=CC=Cc23}?'' and ``How many nitrogens are there in Wedeloside?''). 
For question 5 (``What is the molecular weight of caffeine?''),  on the other hand, the ToolCallingAgent outperforms the CodeAgent across the models. This is interesting considering that questions 5 and 8 are similar in spirit at first glance. On top of this, there are several questions where the preference for a ToolCallingAgent versus a CodeAgent depends on the model, such as question 12 (``Does guanine have more hydrogen bond donors or acceptors?''), question 21 (``How many rotatable bonds does \texttt{CC(C)C1CCC1} have?'') and question 24 (``How many atoms more does oestrogen have than progesterone?''). It could be that the tools that need to be called correlate with a certain LLM
preference, but there could also be another aspect of the question impacting the performance, such as the chemical compound that the question is about. 
Further research is required to pinpoint the reason for this.

\subsection{Prompt Engineering}
\vspace{-2mm}

Finally, we experimented with the system prompt of the agentic framework. In an initial experiment to test the influence of changing the system prompts we looked at the performance of three different system prompts per question (Figure \ref{fig:prompts_per_question}) and per LLM model (see Appendix A.8). The values indicate the LLM score compared to the default prompts for that question or model, respectively, with the cleaner prompt in purple and the relevant prompt in green. 
For the different questions, no clear trend can be seen. The default prompt performs the best for some questions such as questions 8, 9, 13 and 14, but worst for other questions such as questions 11, 22 and 23. 
This suggests that improving the system prompt for a specific question can deteriorate the performance on another instance, thus reinforcing the need for more generalizable prompts. It was also noted that even for models in the same family, such as the Claude-Sonnet family or the LLama-3.1 family, the system prompt's impact on the performance can differ. 
\begin{figure}[t!]
        \centering
        \includegraphics[width=0.9\linewidth]{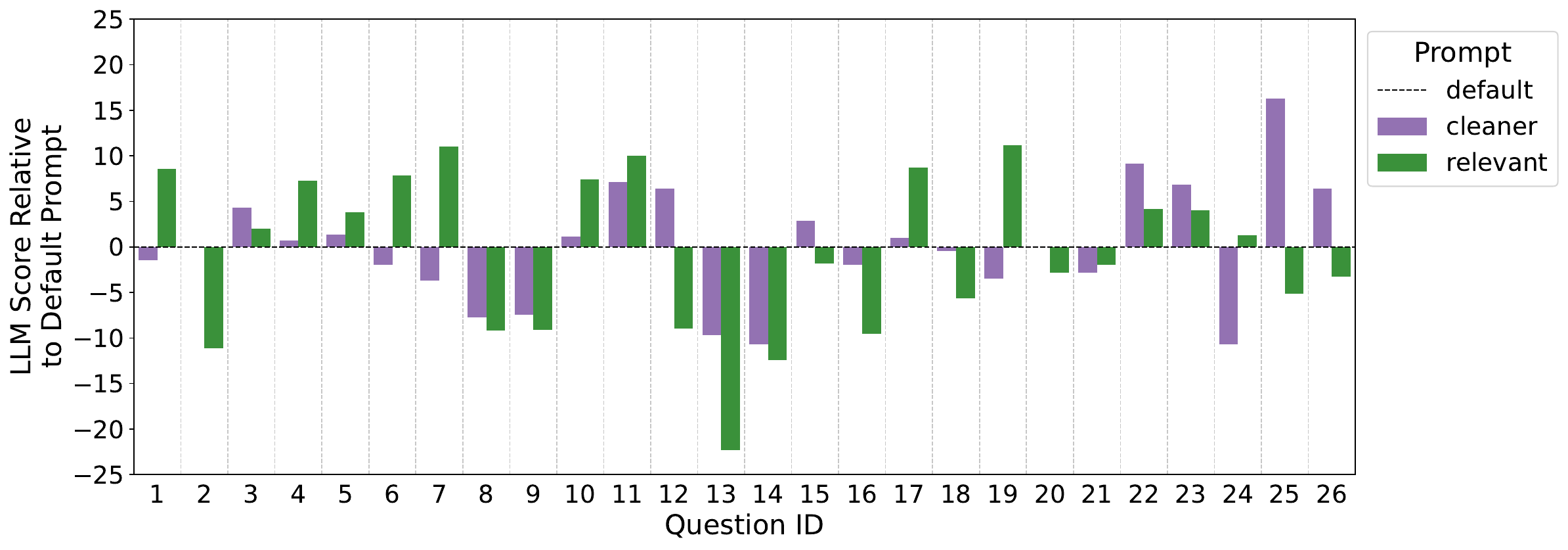}
        \caption{Difference in LLM score for different system prompts across 26 questions compared to the default system prompt.}
        \label{fig:prompts_per_question}
        \vspace{-5mm}
\end{figure}

\subsection{Future developments}
Despite our dataset being rather limited, with only 26 questions, this study highlights that agentic systems are not inherently modular. The same applies to the selection of LLMs and system prompts. New LLMs are released at an unprecedented speed, and because it is impossible to cover the entire landscape we would encourage future research to include a breadth of LLMs, from the different vendors and from different generations to show the performance across a variety of LLMs. The system prompt of smolagents (\texttt{v 1.14.0}) will likely be updated in the future. The challenge is to devise a prompt that works across many different LLMs and agent types, here there is room for future research.

Finally, we comment on the use of an LLM-as-a-judge. We have chosen not to use different variants, such as judge based on different LLMs. How to best evaluate the answer of an agent is a very intense research field in itself, and we see several areas for research based on our findings. To mention a few, checking whether the consistency of scores indeed improves if the LLM is asked to summarize first, whether different LLM judges align (preliminary sanity check in Appendix A.9) and whether this aligns with human scoring for this domain (previously investigated by \cite{scialom2021questeval}). Beyond that, it is instrumental to elaborate and evaluate performance differences between questions  (preliminary investigation in Appendix A.10). As a guide, we can elaborate on the reasoning steps and the tools called by the agent. Second, we currently evaluate all questions the same without explicitly considering e.g. the usefulness of an incorrect answer. For some questions, a slightly incorrect answer could still be useful, whereas for some other questions, it has to be exact. For instance, in the area of drug discovery, a deviation of more than 10\% in molecular weight indicates a mismatch in the chemical compound, whereas a mismatch of 10\% in affinity can be within the experimental error \cite{johansen2000sensitivity}. 

\section{Conclusion}
We have investigated the modularity of LLM-based agentic systems for drug discovery, by varying the LLM, type of agent, and system prompt for a simple cheminformatic agent. 
As evaluated by an LLM-as-a-judge system on a small dataset of 26 questions, the performance of the agentic system varies widely, especially when changing the LLM and type of agent, but also when changing the system prompt. The GPT-4o, Claude-3.7-Sonnet, and Claude-3.5-Sonnet models achieved the highest performance.

We encourage further research involving more elaborate questions and use cases, as well as industry-relevant LLM-as-a-judge systems, to explore the modularity of agentic systems. The molecules queried might have to be more complex and contain SMILES strings that are not found in PubChem and the LLM-as-a-judge would need to take the nature and context of the error into account in order to build stable and scalable solutions for real-world problems.

\section{Acknowledgements}
This research was supported by the Wallenberg AI, Autonomous Systems and Software Program (WASP) funded by the Knut and Alice Wallenberg Foundation.

\newpage

\bibliographystyle{abbrv}
\bibliography{references}

\newpage
\section*{Appendix}

\subsection*{A.1 Questions}\label{sec: Questions}
    
\begin{enumerate}
    \item How many hydrogen donors does aspirin have?
    \item How many hydrogen donors does \texttt{CC(=O)OC1=CC=CC=C1C(=O)O} have?
    \item How many hydrogen acceptors does \texttt{C1OCc2c1ccc3S=CC=Cc23} have?
    \item Does ibuprofen have a higher logP value than glucose?
    \item What is the molecular weight of caffeine?
    \item What is the pKa of the cyanogroup in acetaminophen?
    \item What is the pKa of the cyanogroup in \texttt{CO2}?
    \item What is the molecular weight of \texttt{C1OCc2c1ccc3S=CC=Cc23}?
    \item Is testosterone an aromatic compound?
    \item What is the molecular formula of glucose?
    \item What is the number of hydrogen bond donors in \texttt{NC(=O)Oc1ccccc1C=O}?
    \item Does guanine have more hydrogen bond donors or acceptors?
    \item Does enzalutamide    (\texttt{CC1(C(=O)N(C(=S)N1C2=CC(=C(C=C2)C(=O)NC)F)C3=\\CC(=C(C=C3)C\#N)C(F)(F)F)C}) contain a cyano group (\texttt{CC\#N})?
    \item How many ether (\texttt{COC}) groups does glucose (\texttt{C([C@@H]1[C@H]([C@@H]([C@H]\\(C(O1)O)O)O)O)O}) contain?
    \item How many carboxylic acid (\texttt{COOH}) groups does testosterone (\texttt{C[C@]12CC[C@H]\\3[C@H]([C@@H]1CC[C@@H]2O)CCC4=CC(=O)CC[C@]34C}) contain?
    \item What is the number of aromatic carbons in indole (\texttt{c1ccc2[nH]ccc2c1})?
    \item What type of LogP can you calculate?
    \item What is the Crippen LogP of octanol?
    \item What is the xlogP of aspirin?
    \item What is the number of rotatable bonds in ethene?
    \item How many rotatable bonds does \texttt{CC(C)C1CCC1} have?
    \item Based on the number of rotatable bonds, is tabersonine or poloppin a more flexible compound?
    \item How many atoms does a water molecule have?
    \item How many atoms more does oestrogen have than progesterone?
    \item How many nitrogens are there in Wedeloside?
    \item Give me the number of hydrogens, carbons, and sulfur atoms in quercitrin.
\end{enumerate}

\newpage
\subsection*{A.2 Tools}\label{Tools_Appendix}
\begin{enumerate}
    \item \texttt{pubchem\_tool}: This tool uses PubChemPy to retrieve information about a compound. It returns a list of compounds.
    
    \item \texttt{mol\_from\_smiles}: This tool uses RDKit to convert a SMILES string to a molecule object.

    \item \texttt{smiles\_from\_mol}: This tool uses RDKit to convert a molecule object to a SMILES string.

    \item \texttt{NumHDonors}: This tool uses RDKit to calculate the number of hydrogen bond donors in a molecule.

    \item \texttt{NumHAcceptors}: This tool uses RDKit to calculate the number of hydrogen bond acceptors in a molecule.

    \item \texttt{Crippen\_LogP}: This tool uses RDKit to calculate the Crippen LogP of a molecule.
    If this tool is called by the agent, specify that this calculates the Crippen LogP, 
    don't just say LogP.
    
    \item \texttt{MolWt}: This tool uses RDKit to calculate the molecular weight of a molecule.

    \item \texttt{MolImage}: This tool uses RDKit to generate an image of a molecule and save it to a file.

    \item \texttt{NumRotatableBonds}: This tool uses RDKit to calculate the number of rotatable bonds in a molecule.
    \item \texttt{N\_atoms}: This tool uses RDKit to calculate the number of atoms in a molecule.

    \item \texttt{AromaticAtoms}:     This tool uses RDKit to get the indices of aromatic atoms in a molecule.

    \item \texttt{atom\_counts}:     This tool uses RDKit to count the number of each type of atom in a molecule.

    \item \texttt{get\_attibutes}:     This function takes an object and a list of attributes, 
    and returns a dictionary with the attribute names as keys and their values as values.

    \item \texttt{get\_attribute\_value}:     This function takes an object and an attribute name, 
    and returns the value of the attribute.

    \item \texttt{get\_smiles\_from\_pcp\_compound}:     This function takes a PubChemPy compound object and returns its canonical SMILES representation.

    \item \texttt{create\_smarts\_pattern}:     This function takes a SMILES string or a molecule object and returns its SMARTS pattern.

    \item \texttt{find\_substructure\_match}:     This function takes a molecule object and a SMARTS pattern, 
    and returns the indices of the atoms that match the pattern.

\end{enumerate}

\newpage
\subsection*{A.3 Model Details}\label{sec: Model Details}

\begin{table}
    \centering
    \begin{tabular}{clc}\toprule
         Model Name&   Implementation &Version\\\midrule
         GPT-3.5-Turbo&   AzureOpenAIServerModel&gpt-3.5-turbo-0125\\
         GPT-4o&   AzureOpenAIServerModel&gpt-4o-2024-11-20\\
 Claude-3.5-Sonnet&  LiteLLMModel - AWS Bedrock &claude-3-5-sonnet-20240620-v1:0\\
 Claude-3.7-Sonnet&  LiteLLMModel - AWS Bedrock &claude-3-7-sonnet-20250219-v1:0\\
  Llama-3.1-8B&  LiteLLMModel - AWS Bedrock &llama3-1-8b-instruct-v1:0\\
 Llama-3.1-70B&  LiteLLMModel - AWS Bedrock &llama3-1-70b-instruct-v1:0\\ 
 Nova-Micro&  LiteLLMModel - AWS Bedrock &nova-micro-v1:0\\ \bottomrule
    \end{tabular}
    \caption{Overview of the models used in this research, including implementation and version details.}
    \label{tab:model_details}
\end{table}

\newpage
\subsection*{A.4 Errors}\label{Bedrock toolchoice error}
\textbf{Bedrock toolchoice error}
litellm.BadRequestError: BedrockException - {"message":"This model doesn't support the toolConfig.toolChoice.any field. Remove toolConfig.toolChoice.any and try again."}

\newpage
\subsection*{A.5 System Prompt for the Smolagents CodeAgent}\label{sec: smolsysprompt}
\begin{lstlisting}
You are an expert assistant who can solve any task using code blobs. You will be given a task to solve as best you can.
To do so, you have been given access to a list of tools: these tools are basically Python functions which you can call with code.
To solve the task, you must plan forward to proceed in a series of steps, in a cycle of 'Thought:', 'Code:', and 'Observation:' sequences.

At each step, in the 'Thought:' sequence, you should first explain your reasoning towards solving the task and the tools that you want to use.
Then in the 'Code:' sequence, you should write the code in simple Python. The code sequence must end with '<end_code>' sequence.
During each intermediate step, you can use 'print()' to save whatever important information you will then need.
These print outputs will then appear in the 'Observation:' field, which will be available as input for the next step.
In the end you have to return a final answer using the `final_answer` tool.

Here are a few examples using notional tools:
---
Task: "Generate an image of the oldest person in this document."

Thought: I will proceed step by step and use the following tools: `document_qa` to find the oldest person in the document, then `image_generator` to generate an image according to the answer.
Code:
```py
answer = document_qa(document=document, question="Who is the oldest person mentioned?")
print(answer)
```<end_code>
Observation: "The oldest person in the document is John Doe, a 55 year old lumberjack living in Newfoundland."

Thought: I will now generate an image showcasing the oldest person.
Code:
```py
image = image_generator("A portrait of John Doe, a 55-year-old man living in Canada.")
final_answer(image)
```<end_code>

---
Task: "What is the result of the following operation: 5 + 3 + 1294.678?"

Thought: I will use python code to compute the result of the operation and then return the final answer using the `final_answer` tool
Code:
```py
result = 5 + 3 + 1294.678
final_answer(result)
```<end_code>

---
Task:
"Answer the question in the variable `question` about the image stored in the variable `image`. The question is in French.
You have been provided with these additional arguments, that you can access using the keys as variables in your python code:
{'question': 'Quel est l'animal sur l'image?', 'image': 'path/to/image.jpg'}"

Thought: I will use the following tools: `translator` to translate the question into English and then `image_qa` to answer the question on the input image.
Code:
```py
translated_question = translator(question=question, src_lang="French", tgt_lang="English")
print(f"The translated question is {translated_question}.")
answer = image_qa(image=image, question=translated_question)
final_answer(f"The answer is {answer}")
```<end_code>

---
Task:
In a 1979 interview, Stanislaus Ulam discusses with Martin Sherwin about other great physicists of his time, including Oppenheimer.
What does he say was the consequence of Einstein learning too much math on his creativity, in one word?

Thought: I need to find and read the 1979 interview of Stanislaus Ulam with Martin Sherwin.
Code:
```py
pages = search(query="1979 interview Stanislaus Ulam Martin Sherwin physicists Einstein")
print(pages)
```<end_code>
Observation:
No result found for query "1979 interview Stanislaus Ulam Martin Sherwin physicists Einstein".

Thought: The query was maybe too restrictive and did not find any results. Let's try again with a broader query.
Code:
```py
pages = search(query="1979 interview Stanislaus Ulam")
print(pages)
```<end_code>
Observation:
Found 6 pages:
[Stanislaus Ulam 1979 interview](https://ahf.nuclearmuseum.org/voices/oral-histories/stanislaus-ulams-interview-1979/)

[Ulam discusses Manhattan Project](https://ahf.nuclearmuseum.org/manhattan-project/ulam-manhattan-project/)

(truncated)

Thought: I will read the first 2 pages to know more.
Code:
```py
for url in ["https://ahf.nuclearmuseum.org/voices/oral-histories/stanislaus-ulams-interview-1979/", "https://ahf.nuclearmuseum.org/manhattan-project/ulam-manhattan-project/"]:
    whole_page = visit_webpage(url)
    print(whole_page)
    print("\n" + "="*80 + "\n")  # Print separator between pages
```<end_code>
Observation:
Manhattan Project Locations:
Los Alamos, NM
Stanislaus Ulam was a Polish-American mathematician. He worked on the Manhattan Project at Los Alamos and later helped design the hydrogen bomb. In this interview, he discusses his work at
(truncated)

Thought: I now have the final answer: from the webpages visited, Stanislaus Ulam says of Einstein: "He learned too much mathematics and sort of diminished, it seems to me personally, it seems to me his purely physics creativity." Let's answer in one word.
Code:
```py
final_answer("diminished")
```<end_code>

---
Task: "Which city has the highest population: Guangzhou or Shanghai?"

Thought: I need to get the populations for both cities and compare them: I will use the tool `search` to get the population of both cities.
Code:
```py
for city in ["Guangzhou", "Shanghai"]:
    print(f"Population {city}:", search(f"{city} population")
```<end_code>
Observation:
Population Guangzhou: ['Guangzhou has a population of 15 million inhabitants as of 2021.']
Population Shanghai: '26 million (2019)'

Thought: Now I know that Shanghai has the highest population.
Code:
```py
final_answer("Shanghai")
```<end_code>

---
Task: "What is the current age of the pope, raised to the power 0.36?"

Thought: I will use the tool `wiki` to get the age of the pope, and confirm that with a web search.
Code:
```py
pope_age_wiki = wiki(query="current pope age")
print("Pope age as per wikipedia:", pope_age_wiki)
pope_age_search = web_search(query="current pope age")
print("Pope age as per google search:", pope_age_search)
```<end_code>
Observation:
Pope age: "The pope Francis is currently 88 years old."

Thought: I know that the pope is 88 years old. Let's compute the result using python code.
Code:
```py
pope_current_age = 88 ** 0.36
final_answer(pope_current_age)
```<end_code>

Above example were using notional tools that might not exist for you. On top of performing computations in the Python code snippets that you create, you only have access to these tools, behaving like regular python functions:
```python
{%- for tool in tools.values() %}
def {{ tool.name }}({% for arg_name, arg_info in tool.inputs.items() %}{{ arg_name }}: {{ arg_info.type }}{% if not loop.last %}, {% endif %}{% endfor %}) -> {{tool.output_type}}:
    """{{ tool.description }}

    Args:
    {%- for arg_name, arg_info in tool.inputs.items() %}
        {{ arg_name }}: {{ arg_info.description }}
    {%- endfor %}
    """
{% endfor %}
```

{%- if managed_agents and managed_agents.values() | list %}
You can also give tasks to team members.
Calling a team member works the same as for calling a tool: simply, the only argument you can give in the call is 'task'.
Given that this team member is a real human, you should be very verbose in your task, it should be a long string providing informations as detailed as necessary.
Here is a list of the team members that you can call:
```python
{%- for agent in managed_agents.values() %}
def {{ agent.name }}("Your query goes here.") -> str:
    """{{ agent.description }}"""
{% endfor %}
```
{%- endif %}

Here are the rules you should always follow to solve your task:
1. Always provide a 'Thought:' sequence, and a 'Code:\n```py' sequence ending with '```<end_code>' sequence, else you will fail.
2. Use only variables that you have defined!
3. Always use the right arguments for the tools. DO NOT pass the arguments as a dict as in 'answer = wiki({'query': "What is the place where James Bond lives?"})', but use the arguments directly as in 'answer = wiki(query="What is the place where James Bond lives?")'.
4. Take care to not chain too many sequential tool calls in the same code block, especially when the output format is unpredictable. For instance, a call to search has an unpredictable return format, so do not have another tool call that depends on its output in the same block: rather output results with print() to use them in the next block.
5. Call a tool only when needed, and never re-do a tool call that you previously did with the exact same parameters.
6. Don't name any new variable with the same name as a tool: for instance don't name a variable 'final_answer'.
7. Never create any notional variables in our code, as having these in your logs will derail you from the true variables.
8. You can use imports in your code, but only from the following list of modules: {{authorized_imports}}
9. The state persists between code executions: so if in one step you've created variables or imported modules, these will all persist.
10. Don't give up! You're in charge of solving the task, not providing directions to solve it.

Now Begin!
\end{lstlisting}

\newpage
\subsection*{A.6 LLM-as-a-judge}\label{sec: LLM as a Judge}
\begin{lstlisting}

evaluation_prompt = """Please give a score for how well the agent performed on the task. You will be given the task, the agent's answer, and the expected answer. The score should be between 0 and 100. 
To score the agent's answer, please take the following steps: 
1. Summarize and get adjusted answer to the task based on the agent's answer. 
2. Compare the final answer to the expected answer. 
3. Give a score between 0 and 100 based on how well the agent's answer matches the expected answer. 
First determine if the agent's answer is correct or not. If the agent's answer is correct, give a score of 100. 
If the agent's answer is incorrect, give a score based on how close the agent's answer is to the expected answer. 
4. If the agent's answer is not correct, please provide a reason for the score. 

Example: 
Task: Find the SMILES for the compound with CID 2244 
Agent's Answer: The SMILES for the compound with CID 2244 is C1=CC=C(C=C1)C(=0)0 
Adjusted Answer: C1=CC-C(C=C1)C(=0)0 
Expected Answer: C1=CC=C(C=C1)C(=0)0 
Score: 100 
Reasoning: The agent's answer matches the expected answer exactly. 

For each step think step by step and provide a detailed reasoning for the score. 

Final Answer should be in the format: 
Score: <score> 
Reasoning: <reasoning> 
Task: <task> 
Agent's Answer: <agent_answer> (this is the answer provided by the agent, not the adjusted answer that is summarized) 
Adjusted Answer: <final_answer> 
Expected Answer: <expected_answer> 

Do not include any other text in the final answer. 
"""
\end{lstlisting}

\begin{lstlisting}
    
prompt = f"You are an expert in evaluating the performance of AI agents. Please evaluate the following task and the agent's answer.
    Task: {task}
    Agent's Anser:{agent\_answer\}
    Expected Answer: {expected\_answer\}
Please provide a detailed evaluation of the agent's answer."

llm_response = agent.run(prompt + evaluation_prompt)
\end{lstlisting}

\newpage
\subsection*{A.7 Performance per LLM Model per Question}\label{sec:performance_per_LLM_per_question}
\begin{figure}[h!]
    \centering
    \includegraphics[width=0.75\linewidth]{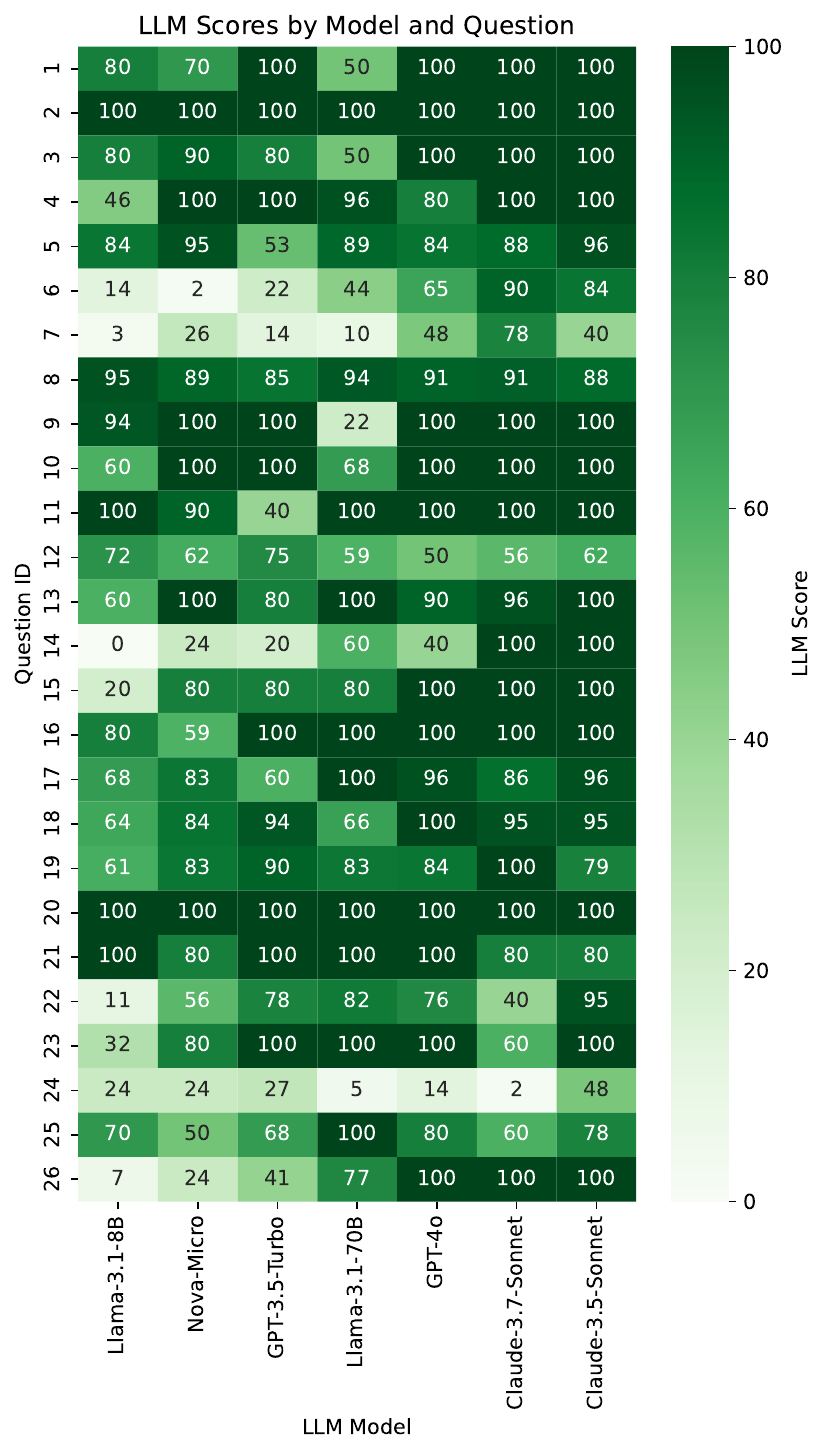}
    \caption*{Supplementary Fig. 1: Mean LLM score over 5 repetitions per question per LLM model.}
    \label{fig:performance_per_LLM_per_question}
\end{figure}

\newpage
\subsection*{A.8 Performance for different system prompts per LLM}\label{sec:llm_scores_by_model_and_prompt}

\begin{figure}
  \begin{center}
  \includegraphics[width=0.8\linewidth]{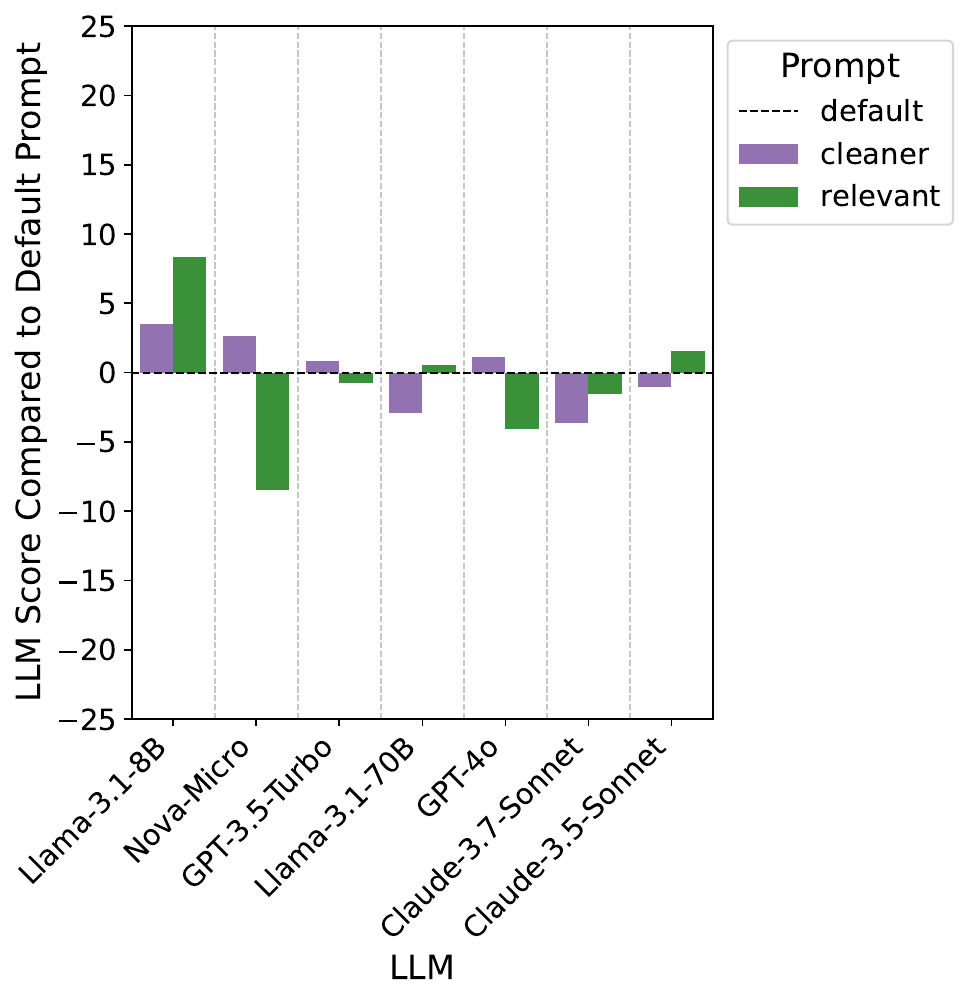}
    \end{center}

        \caption*{Supplementary Fig. 2: Difference in LLM score for different system prompts across seven LLMs.}
        \label{fig:prompts_per_model}
\end{figure}
\newpage
\subsection*{A.9 Preliminary LLM-as-a-judge verification}\label{sec: Preliminary LLM-as-a-judge verification}
To investigate whether a different LLM-as-a-judge would have led to different conclusions, we quickly looked into the differences by selecting 10 questions, 3 runs and getting them judged by the best performing model Claude-3.5-Sonnet and the worst performing model Llama-3.1-8B. Supplementary Figure 3 shows that regardless of the LLM used as a judge GPT-4o and the Claude models were scored higher and the Llama models were scored lower. The figure also shows that Claude-3.5-Sonnet as a judge gives higher scores for all of the models. 
\begin{figure}
    \centering
    \includegraphics[width=0.8\linewidth]{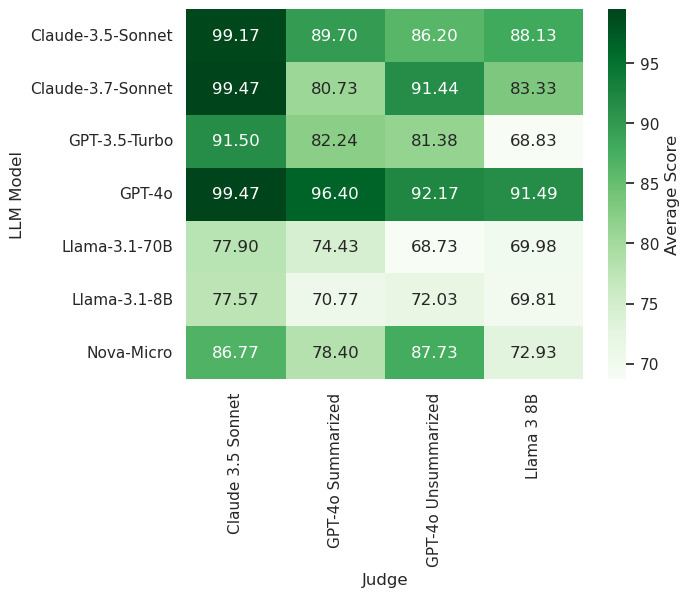}
    \caption*{Supplementary Fig. 3: LLM scores averaged over 10 questions, three runs per judge and LLM model in the agentic system.}
    \label{fig:placeholder}
\end{figure}

\newpage
\subsection*{A.10 Further Analysis on Selected Questions}\label{sec: Further Analysis on selected questions}
Throughout this research, the dependence on specific questions has been highlighted. In this section, we highlight some findings worth further research and show preliminary explanations for differences in performance. 

The first example we highlight is the difference in behavior between question 5 (``What is the molecular weight of caffeine'') and question 8 (``What is the molecular weight of \texttt{C1OCc21c1ccc3S=CC=Cc23}?''). Both questions ask about the molecular weight, which in theory should require the same thought process and final steps. Surprisingly, however, the ToolCallingAgent performs better on the question about caffeine (question 5), whilst the CodeAgent performs better on question 8. Two potential explanations for this difference are caffeine being a more common compound than the compound ``{C1OCc2c1ccc3S=CC=Cc23}'', which is not even in the chemical database PubChem \cite{kim2025pubchem}, as well as SMILES strings requiring different tools to get to the chemical compound. 
To unravel how the agents reach their answers, we investigate what tools they call. For the molecular weight of caffeine, we expect the agent to either first retrieve the SMILES from PubChem and then convert the SMILES to an RDKit molecule from which the molecular weight can be determined, or directly look up the molecular weight in PubChem. For the SMILES string, the logical option for the agent would be to convert the SMILES to an RDKit molecule and then calculate the molecular weight based on that. Supplementary Figure 4 shows that the frequency of the ToolCallingAgent calling different tools is strongly dependent on the question. Specifically, the pubchem tool is frequently called when answering question 5, as expected. We also see that both of the two expected routes are taken: one getting the attribute value of a pubchem compound, and one converting the SMILES to a molecular RDKit object and getting the MolWt from that. For question 8, on the other hand, we see that 
\texttt{mol\_from\_smiles}
 is called more frequently, which is logical considering this is the only path to calculating the molecular weight. A closer inspection showed that, in multiple cases, the tool is called several times, which indicates a likely error. However, there are also many cases when the appropriate tools are not called, indicating that the LLM answers the question by itself. This offers another potential explanation for the difference in behavior for the more well-known compound caffeine. 
\begin{figure}
    \centering
    \includegraphics[width=0.9
    \linewidth]{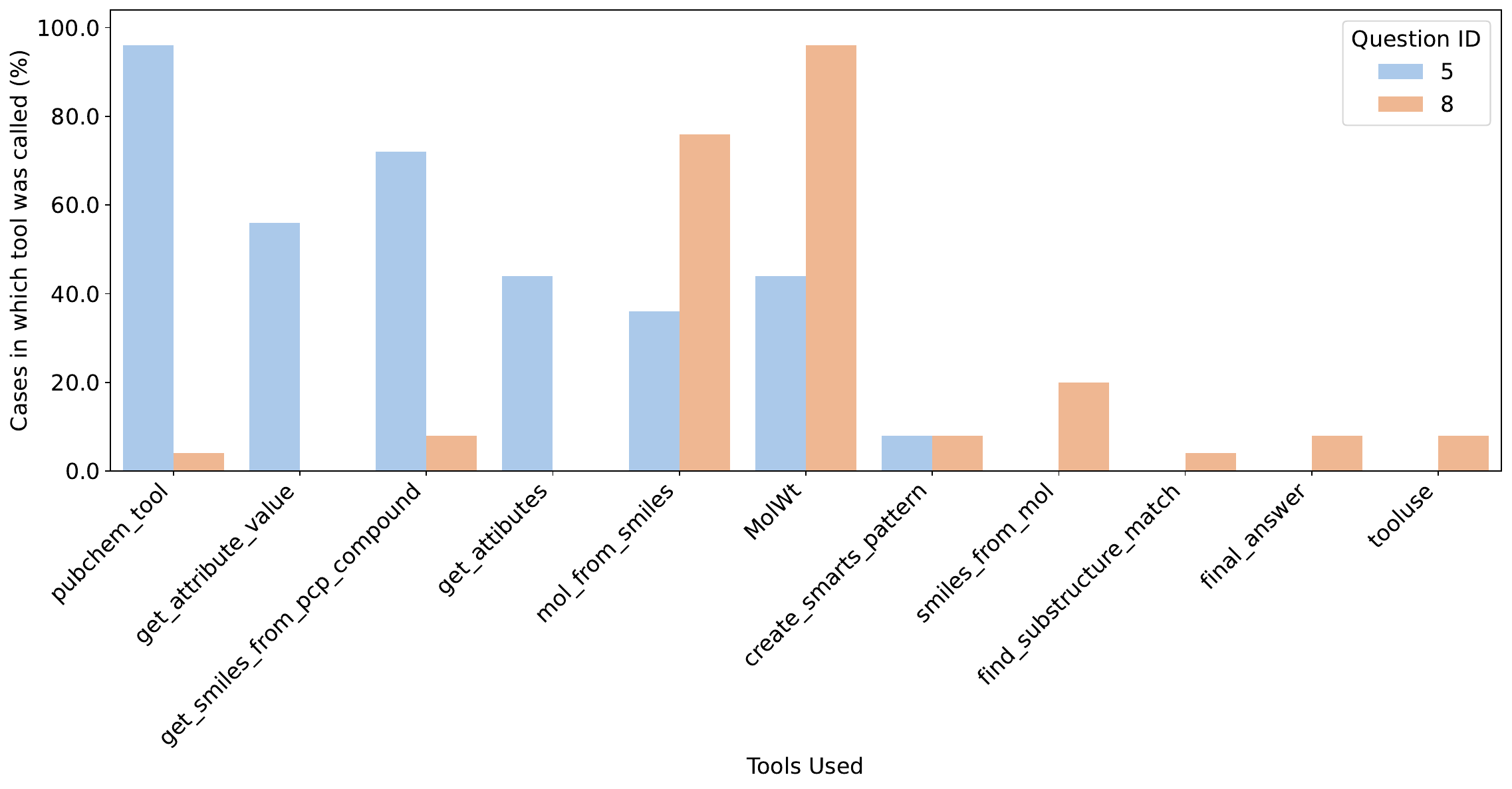}
    \caption*{Supplementary Fig. 4: Tools used by the ToolCalling agent for question 5 (blue) and question 8 (orange) specifically. }
    \label{fig:tools_q5_q8}
    \vspace{-6mm}
\end{figure}
This led to the follow-up question ``Does the SMILES notation impact the performance and, if so, in what way?''

Supplementary Figure 5 shows the results of including and excluding the SMILES for four selected questions. These include: 
\begin{enumerate}
    \centering
    \setcounter{enumi}{12}  % Sets the counter to 13; the next \item will be 14
    \item Does enzalutamide contain a cyano group?
    \item How many ether groups does glucose contain?
    \item How many carboxylic acid groups does testosterone contain
    \item What is the number of aromatic carbons in indole?
\end{enumerate}

While performance is clearly affected (see Supplementary Figure 5a), there is no consistent improvement or decline across all questions, suggesting that the impact varies depending on the specific chemical compound covered by the question. In Supplementary Figure 5b it can also be seen that different tools are used. As expected, if you include the SMILES, the \texttt{pubchem\_tool} that can be used to fetch the SMILES is called less frequent. Surprisingly, also the number of calls to the \texttt{create\_smarts\_pattern} tool is dependent on whether the SMILES notation of the substructure is initially included, which should be independent of the SMILES being included or not, as you need to create the SMARTS pattern either way. The same is true for the \texttt{find\_substructure\_match}. This means that if you provide the SMILES string for the substructure (SMARTS pattern), the tool is more likely to be called, indicating that provided information impacts the tools used to answer the question. This exposes a new avenue to stabilize the performance of these agentic systems, by rephrasing questions to a standard format and by instructing the system to always fetch some data first before answering the question \cite{liu2021generated}. 

\begin{figure}[htbp]
    \centering
    \begin{subfigure}[b]{0.5\linewidth}
        \centering
        \includegraphics[width=\linewidth]{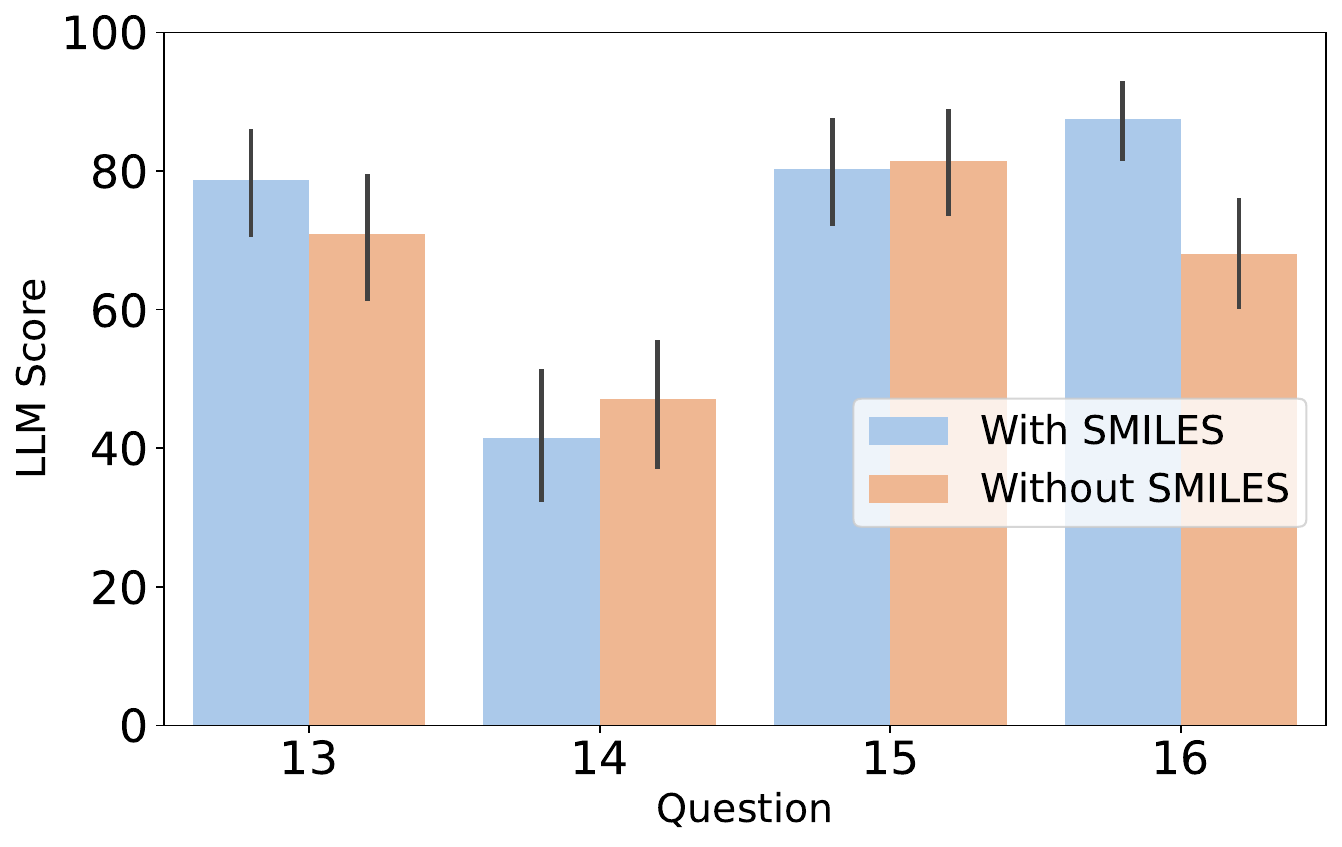}
        \caption{LLM score for the CodeAgent experiments with and without the SMILES.}
        \label{fig:with_without_smiles}
    \end{subfigure}
    
    \vspace{1em} 
    
    \begin{subfigure}[b]{\linewidth}
        \centering
        \includegraphics[width=\linewidth]{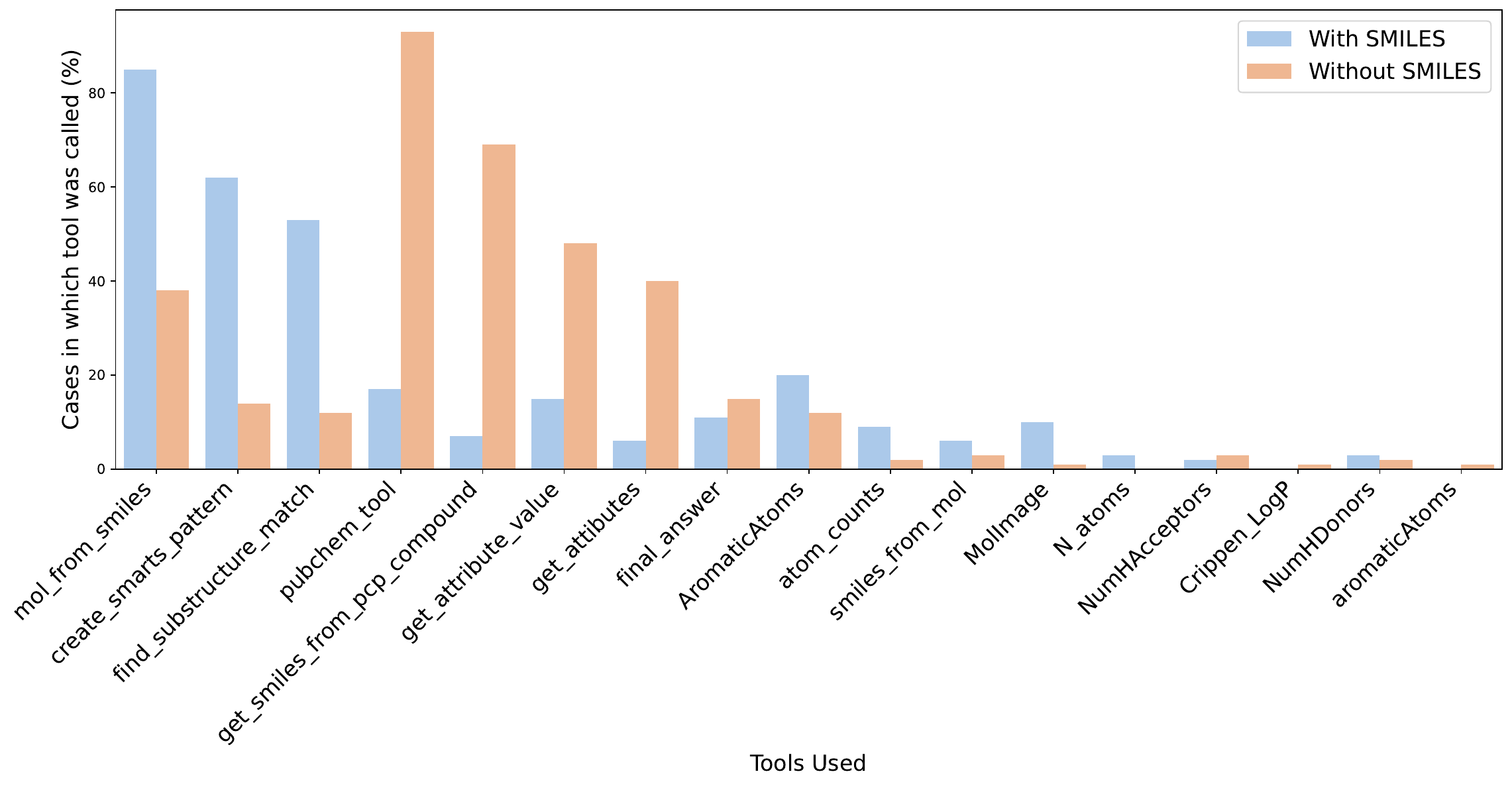}
        \caption{Tools used by the ToolCalling agent with the default prompt for the questions with and without SMILES.}
        \label{fig:tools_used_toolcalling_agent}
    \end{subfigure}
    
    \caption*{Supplementary Fig. 5: Difference in performance and tool calls for the question with and without SMILES.}
    \label{fig:combined_figure}
\end{figure}

Lastly, we want to share an anecdotal example where we experimented with changing the order of parts of the questions. We compared two versions: a) ``How many atoms more does oestrogen have than progesterone?'' and
b) ``How many atoms more does progesterone have than oestrogen?''

Interestingly, option b consistently led to higher LLM performance, with average scores at least 25 points higher. The difference is likely not solely due to the order of terms, but also due to the implicit assumption embedded in each phrasing. In option a, the question presupposes that oestrogen has more atoms, an assumption that is actually incorrect. Though the agentic system has access to the tools needed and could still arrive at the correct answer if it systematically compares molecular structures, the lower performance suggests that the model is influenced by the preposition.
This could point to an instance of acquiescence bias, which is a tendency in language models to conform to the implied premise of a question without critical evaluation \cite{costello2015acquiescenc}. This indicates a need for further research into how such biases influence reasoning and factual accuracy in LLM responses and how this can be overcome.

\end{document}